\documentclass[letterpaper]{article} 
\usepackage{aaai23}  
\usepackage{times}  
\usepackage{helvet}  
\usepackage{courier}  
\usepackage[hyphens]{url}  
\usepackage{graphicx} 
\usepackage{natbib}  
\usepackage{caption} 
\usepackage{amsfonts}
\usepackage{multirow}
\usepackage{algorithm}
\usepackage{algorithmic}

\usepackage{newfloat}
\usepackage{listings}
\DeclareCaptionStyle{ruled}{labelfont=normalfont,labelsep=colon,strut=off} 
\floatstyle{ruled}
\newfloat{listing}{tb}{lst}{}
\floatname{listing}{Listing}
\title{Fair Multi-Exit Framework for Facial Attribute Classification}
\author{
    Ching-Hao Chiu\textsuperscript{\rm 1}\equalcontrib,
    Hao-Wei Chung\textsuperscript{\rm 1}\equalcontrib, 
    Yu-Jen Chen\textsuperscript{\rm 1},
    Yiyu Shi \textsuperscript{\rm 2},
    Tsung-Yi Ho\textsuperscript{\rm 1}
}
\affiliations{
    \textsuperscript{\rm 1} Department of Computer Science, National Tsing Hua Unviserity\\
    \textsuperscript{\rm 2} Department of Computer Science and Engineering, University of Notre Dame\\
    { {\{gwjh101708, xdmanwww, yujenchen\}@gapp.nthu.edu.tw}}\\
    yshi4@nd.edu,
    tyho@cs.nthu.edu.tw

}

\usepackage{bibentry}

\begin{document}

\maketitle

\begin{abstract}
Fairness has become increasingly pivotal in facial recognition. Without bias mitigation, deploying unfair AI would harm the interest of the underprivileged population. In this paper, we observe that though the higher accuracy that features from the deeper layer of a neural networks generally offer, fairness conditions deteriorate as we extract features from deeper layers. This phenomenon motivates us to extend the concept of multi-exit framework. Unlike existing works mainly focusing on accuracy, our multi-exit framework is fairness-oriented, where the internal classifiers are trained to be more accurate and fairer. During inference, any instance with high confidence from an internal classifier is allowed to exit early. Moreover, our framework can be applied to most existing fairness-aware frameworks. Experiment results show that the proposed framework can largely improve the fairness condition over the state-of-the-art in CelebA and UTK Face datasets.  
\end{abstract}

\section{Introduction}
\label{sec:intro}

Machine learning has been applied in various fields and has impacted our daily life in recent years. Many institutions have introduced machine learning-based systems to help them decide on administrative operations. Although the machine learning model achieves accurate prediction, there exists some bias in such an AI system \cite{mehrabi2021survey, dressel2018accuracy}. The discriminative nature of the machine learning model will harm the opportunity of different races, religions, and genders and thus tear society apart.

Several methods are proposed to ameliorate the bias in machine learning models. Many of them \cite{wang2022fairness, zhang2018mitigating, kim2019learning, ngxande2020bias} adopted adversarial training to eliminate bias by training the network to learn a classifier while disabling the adversary's ability to categorize the sensitive attribute. Disentanglement representation \cite{creager2019flexibly} is another mainstream to achieve fairness. It forces the latent vector of the sensitive group to be independent of that of the target group and thus reaches fairness.

In this paper, we observe that although features from a deep layer of a neural network bring high accuracy in classification, they cause fairness conditions to deteriorate, and we will demonstrate this observation in Section \ref{sec:motivation}. 
This finding reminds us of the ``overthinking'' phenomenon in deep neural networks \cite{kaya2019shallow}, 
where accuracy decreases as the features come from deeper in a neural network. 
This problem is successfully addressed through multi-exit neural networks by introducing
multiple internal classifiers and treating high-confidence results from these internal 
classifiers as the final result. We conjecture that a similar approach can be used 
to address the issue concerning fairness. 

In the proposed multi-exit framework for fairness, both accuracy and fairness constraints are included when training every internal classifier to keep the fairness and accuracy from shallow to deep. Specifically, with early exits at the inference stage, a sufficient discriminative basis can be obtained based on low-level features when classifying easier samples. This contributes to selecting the optimal prediction for each test instance regarding the trade-off between accuracy and fairness.

To validate the effectiveness of our framework, we perform the facial attribute classification on CelebA  \cite{liu2018large} and UTK Face \cite{zhang2017age} datasets. Experiments show that the proposed method significantly improves the results from the baseline and different state-of-the-arts in terms of the trade-off between classification accuracy and fairness.

The main contributions of the proposed method are as follows: 
\begin{itemize}
    \item Extensive experiments show that although the features from a deep layer of a neural network are highly discriminative and thus bring high accuracy, they cause fairness to deteriorate.
    \item We explore the use of multi-exit training framework to deal with the fairness issue.
    \item We introduce a simple debias framework with high extensibility that could apply to different baseline and state-of-the-art, and achieve further improvement.
    \item Through extensive experiments on different settings and comparisons, our framework achieves the best trade-off performances between top-1 accuracy and fairness on CelebA and UTK Face datasets.

\end{itemize}



\section{Motivation}
\label{sec:motivation}
\begin{figure}[ht]
\begin{center}

\includegraphics[width=0.99\linewidth]{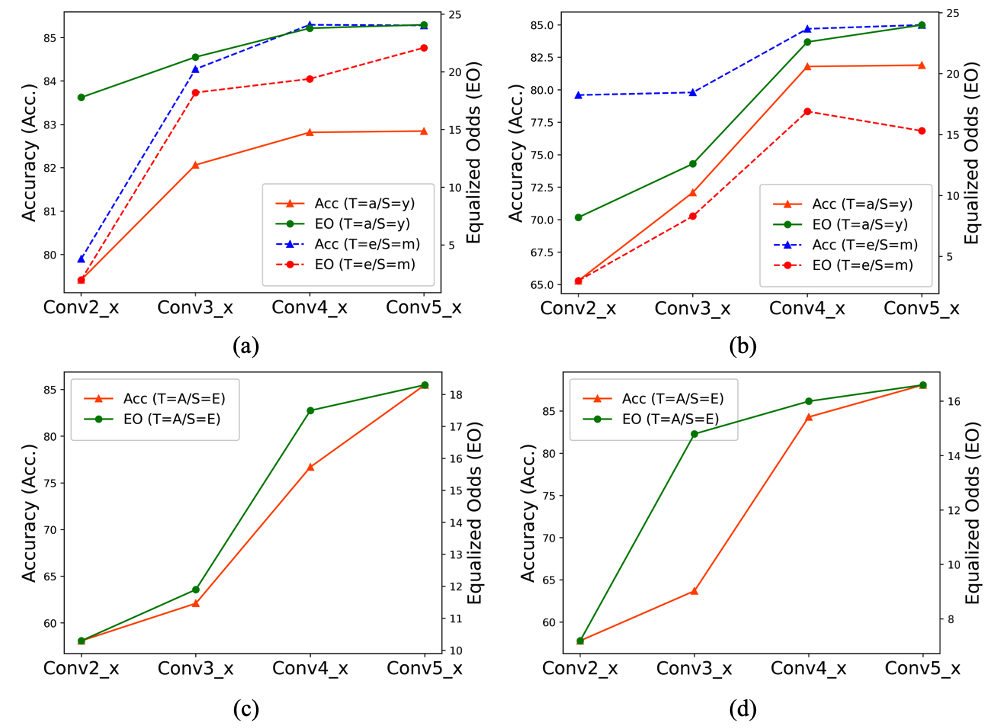}

\caption{Equalized odds (EO) and accuracy (Acc.) for the internal layers of the conventional ResNet on CelebA (a-b) and UTK Face (c-d) dataset. Note that lower EO represents fairer. (a) and (c) are the results using ResNet-18, while (b) and (d) use the ResNet-50. T and S stand for target and sensitive attributes, respectively. In the CelebA dataset, $a$, $e$, $m$, and $y$ represent \emph{attractiveness}, \emph{bags-under-eyes}, \emph{male}, and \emph{young}, respectively. For the UTK Face dataset, $A$ and $E$ are the \emph{Age} and \emph{Ethnicity}, respectively.
}
\label{Fig.Motivation}
\end{center}
\end{figure}

As shown in Fig. \ref{Fig.Motivation}, we observe that despite the features from a deep layer of a neural network bring high accuracy in classification, they cause fairness condition to deteriorate. We report the equalized odds (EO) and accuracy (Acc.) of ResNet-18 and ResNet-50 on the CelebA and the UTK Face dataset. We first trained a vanilla CNN, and froze the backbone. Then, we trained 3 MLP classifiers with the features from each residual module (i.e., Conv2\_x, Conv3\_x, Conv4\_x, Conv5\_x).

For CelebA dataset, in the experiments of the ResNet-18 (Fig. \ref{Fig.Motivation}(a)) and the ResNet-50 (Fig. \ref{Fig.Motivation}(b)), both the equalized odds (EO) and accuracy (Acc.) increase when the features are extracted from deeper layers. When comparing the accuracy between conv2\_x and the final layer in ResNet-18, the final layer has improved 6\% on average. However, the EO also increases over six times larger on average. As for the ResNet-50, the accuracy increased by about 16\%, and the EO increased more than eight times larger on average.

In Fig. \ref{Fig.Motivation}(c) and (d), results on UTK Face dataset again show a similar trend on the increasing EO and accuracy from shallow to deep layers. The ResNet-18 (Fig. \ref{Fig.Motivation}(c)) shows a 47\% improvement in accuracy with 1.7 times higher EO from the shallow layer (conv2\_x) to the deeper layer (Final layer). As for the ResNet-50 (Fig. \ref{Fig.Motivation}(d)), accuracy increases about 52\% and EO increases over 2.3 times.

As higher EO stands for larger bias (unfair) of the predicted result, intuitively, choosing the result at a shallow layer for final prediction could ameliorate the bias condition of the model. Our extensive experiments demonstrate that our observation could be applied to different network architectures and datasets.

\section{Related Work}
\begin{figure*}[t]
\begin{center}

\includegraphics[width=0.85\linewidth]{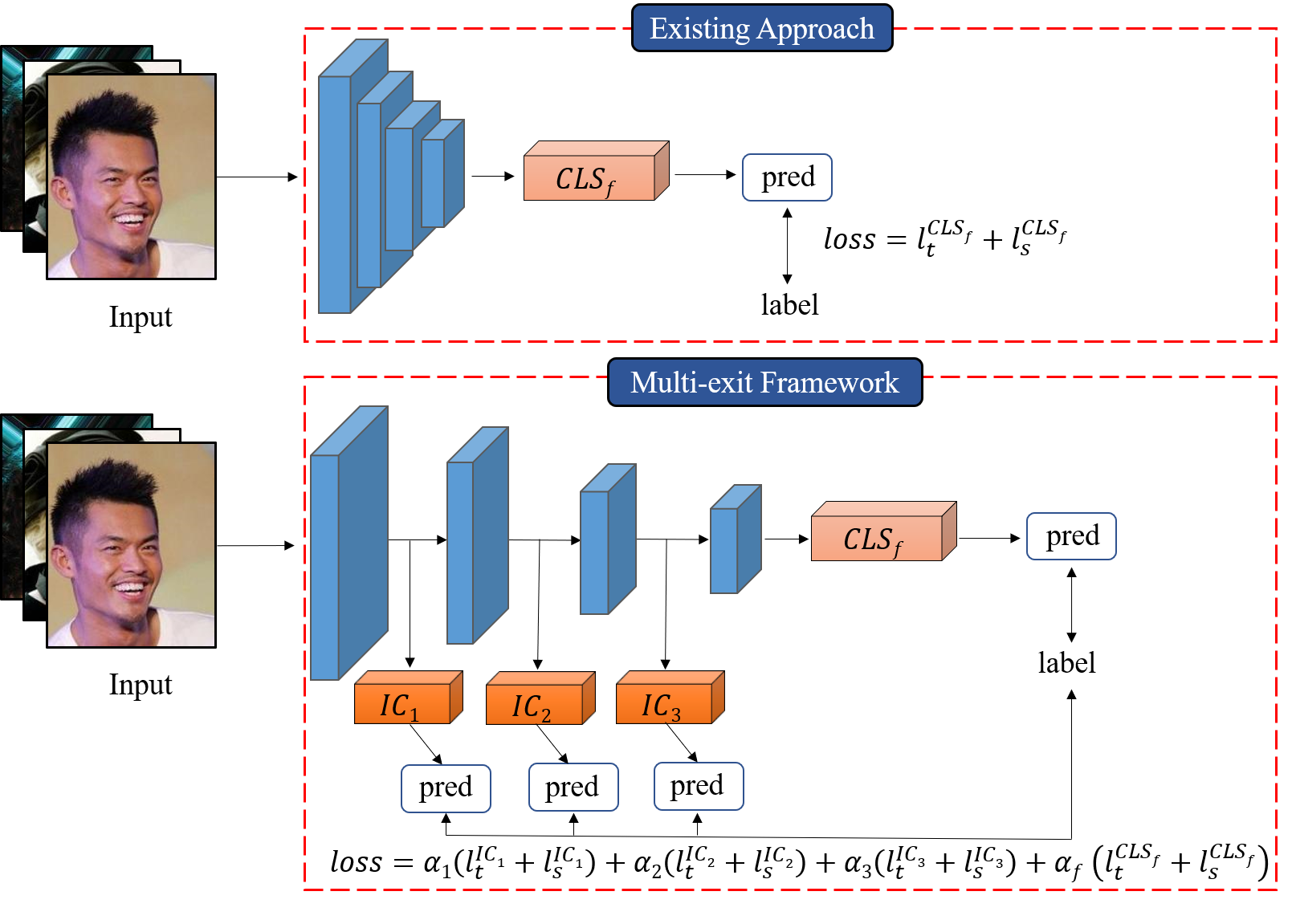}
\caption{Illustration of the multi-exit training framework. $l_t$ and $l_s$ are the loss function related to target and sensitive attributes, respectively.
}
\label{Fig.Method}
\end{center}
\end{figure*}

\subsection{Multi-Exit Networks and Early Exit Policy} 
The multi-exit network is designed by putting additional loss constraints at internal exit branches (internal classifier) to increase the accuracy at shallow layer. With the early exit scheme, early exit branches reduce the computational resource during the model inference time while enhancing the model's accuracy. In the inference phase, the instances will stop inferencing and leave the model from different branches following the pre-defined early exit rules or criteria. BranchyNet \cite{teerapittayanon2016branchynet} calculates entropy at each branch after obtaining the result and exits if the entropy of the predicted result is less than the threshold value. Shallow-Deep Networks \cite{kaya2019shallow} calculates the softmax score of each internal classifier's prediction and takes the maximum probability value as the confidence score. Once the score exceeds the threshold during the forward passing, the instance will exit from the branch prematurely.This model further mitigates the ``overthinking'' problem of deep neural networks. \cite{schwartz2020right} leveraged the early exit inference scheme of \cite{kaya2019shallow} and applied a confidence-based strategy to the natural language processing task. \cite{zhou2020bert} makes predictions using adjacent layers and stops inference when the predicted value of the internal classifier remains constant in a given inference unit times. In this paper, we borrow the confidence-based early exit strategy proposed by \cite{kaya2019shallow}to make sure the the early exit instance is confident enough to be correct. To our best knowledge, we are the first work that leverages the early exit technique to improve fairness.

\subsection{Bias Mitigation Methods}
Bias mitigation methods are designed to reduce the native bias in the dataset to reduce the chance of unfair prediction. There are largely three avenues for current debiased strategy, including pre-processing, in-processing, and post-processing. 
(1) Pre-processing strategies usually remove the information which may cause ``discrimination'' from training data before training. \cite{kamiran2012data} use different weight to neutralize the effect of the sensitive information in training phase and present the experiments result on real-life data. \cite{ngxande2020bias} and \cite{lu2020gender} achieve fairness via data pre-process methods including data generation and data augmentation.

In-processing method usually modify on-the-shelf model architecture, loss function, and model regularization to achieve fairness goal. Adversarial training mitigates the bias through adversarially trains an encoder and classifier to learn a fair representation. \cite{zhang2018mitigating} adversarially cooperate a predictor and an adversary to remove the sensitive attributes from the representation. \cite{kim2019learning} eliminated the correlations between extracted feature and sensitive attribute to achieve fairness by unlearn the bias in the feature domain. \cite{wang2022fairness} adversarially learned a perturb to mask out the sensitive information of the input images, and the proposed framework do not need to alter the deep models' parameters and structure. Some regularization-based methods, such as \cite{quadrianto2019discovering} used Hilbert-Schmidt norm to learn a fair representation that retain the features' semantics from input domain. \cite{jung2021fair} learned a fair representation by distilling the fair information of the teacher model to the student model with the Maximum Mean Discrepancy (MMD) \cite{gretton2012kernel} loss. \cite{park2022fair} introduced a group-wise normalization and penalizing the inclusion of sensitive attribute to mitigate the intrinsic unbiased condition of supervised contrastive learning. 

Post-processing method aims to calibrate the model's output to enhance fairness. They need to use sensitive attribute and prediction distribution to modify the previous distribution result. \cite{hardt2016equality} reveals the limit of demographic parity and give a new metric to fairness, equalized odds, and show how to adjust the learned prediction. \cite{zhao2017men} used Lagrangian relaxation to designed an inference algorithm which reduces bias but maintain accuracy in the meantime.

In this paper, we focus on improving the existed in-processing methods by introducing a general multi-exit (ME) training framework. We compare the regularization-based \cite{kim2019learning, jung2021fair} fairness methods w and w/o the ME framework. Moreover, we show that our framework could apply to complex training structures; for instance, the adversarial debias method \cite{kim2019learning} and the fair contrastive learning \cite{park2022fair}.




\section{Method}


In this section, we provide a clear definition of our goal: overcoming the prediction bias of the deep neural network, and we define our problem formulation in \ref{sec:method_problem}. Afterward, in Section \ref{sec:method_main}, we introduce our main approach, multi-exit (ME) training framework and the early exit policy, which allow us to improve the fairness of state-of-the-arts.

\subsection{Problem Formulation}
\label{sec:method_problem}

In the classification task, define input features $ X = x \in \mathbb{R}^{d}$,
target class $ Y  = y \in\{1,2, ..., N \} $, predicted class $\tilde{Y}=\tilde{y} \in \{1,2, ..., N\}$, and sensitive attributes $A  = a \in \{1,2, ..., M \} $.
The goal is to learn a classifier $f : x \rightarrow y$ that predicts the target class $Y$ to achieve high accuracy while being unbiased to the sensitive attributes $A$. Several criteria are proposed to evaluate the bias against sensitive attributes $A$, and we will discuss the fairness criteria in our experiments in Section \ref{sec:evaluation_metrics}.

\begin{table*}[ht]
\centering
\caption{Classification results of the fairness (EO) and accuracy (Acc.) evaluation on the test set of the CelebA dataset. * denotes our own implementation. }
\setlength{\tabcolsep}{4pt}
\label{Table:Compare_SOTA_CelebA}
\begin{tabular}{c||c|c||c|c||c|c||c|c||c|c||c|c}
\hline
\multirow{2}{*}{Methods}           & \multicolumn{2}{c||}{T=a / S=m} &\multicolumn{2}{c||}{T=a / S=y} &\multicolumn{2}{c||}{T=b / S=m}& \multicolumn{2}{c||}{T=b / S=y}& \multicolumn{2}{c||}{T=e / S=m} & \multicolumn{2}{c}{T=e / S=y}  \\ \cline{2-13}
                  & EO            & Acc.          & EO            & Acc.           & EO           & Acc.          & EO           & Acc.          & EO           & Acc.          & EO           & Acc.          \\ \hline \hline
CNN                   & 27.8          & 79.6          & 16.8          & 79.8           & 17.6         & 84.0          & 14.7         & 84.5          & 15.0         & 83.9          & 12.7         & 83.8          \\ 
\textbf{ME-CNN}       & \textbf{23.7} & \textbf{82.3} & \textbf{16.1} & 76.8           & \textbf{12.9} & \textbf{84.8}         & \textbf{12.9} & \textbf{84.8}         & \textbf{10.8} & 82.5          & \textbf{8.1} & 83.6          \\ \hline
LNL                   & 21.8          & 79.9          & 13.7          & 74.3           & 10.7         & 82.3          & 6.8          & 82.3          & 5.0          & 81.6          & 3.3          & 80.3          \\ 
\textbf{ME-LNL}       & \textbf{14.4} & \textbf{82.2} & \textbf{13.1} & 72.7           & \textbf{7.3} & \textbf{82.8} & \textbf{5.5} & \textbf{83.1} & \textbf{2.7} & \textbf{84.0} & \textbf{1.0} & \textbf{82.6} \\ \hline
HSIC*                 & 19.4          & 81.7          & 16.5          & 80.3           & 11.2         & 80.8          & 10.5         & 82.6          & 12.5         & 84.0          & 7.4          & 84.2          \\ 
\textbf{ME-HSIC}      & \textbf{12.9} & 78.8          & \textbf{15.9} & 78.7           & \textbf{3.6} & \textbf{80.9} & \textbf{3.2} & 82.2          & \textbf{7.8} & 83.5          & \textbf{4.1} & 82.0          \\ \hline
FSCL+                 & 6.5           & 79.1          & 12.4          & 79.1           & 4.7          & 82.9          & 4.8          & 84.1          & 3.0          & 83.4          & 1.6          & 83.5          \\ 
\textbf{ME-FSCL+}     & \textbf{5.8}  & 78.2          & \textbf{7.6}  & 76.4           & \textbf{3.6} & 81.2          & \textbf{2.8} & \textbf{84.4} & \textbf{2.0} & \textbf{83.5} & \textbf{1.4} & 80.8          \\ \hline
MFD                   & 7.4           & 78.0          & 14.9          & 80.0           & 7.3          & 78.0          & 5.4          & 78.0          & 8.7          & 79.0          & 5.2          & 78.0          \\ 
\textbf{ME-MFD}       & \textbf{5.8}  & \textbf{78.3} & \textbf{11.4} & 79.5           & \textbf{2.6} & \textbf{82.1} & \textbf{3.3} & \textbf{82.6} & \textbf{1.4} & \textbf{81.9} & \textbf{1.5} & \textbf{84.2} \\ \hline
\end{tabular}
\end{table*}

\subsection{Multi-Exit (ME) Training Framework}
\label{sec:method_main}

Fundamental to our approach is that although deep neural networks usually achieve high accuracy in the deeper layer, the prediction would be unfair to the different sensitive groups, e.g., race, gender, age, etc. This phenomenon allows the possibility to select the result at a shallow layer with high confidence to solve the unfair issue and maintain the predicted accuracy. Our method is based on a multi-exit training framework and an early exit policy similar to previous works \cite{kaya2019shallow}. The main contribution lies in introducing the use of multi-exit to improve the fairness of most state-of-the-arts.

As shown in Fig. \ref{Fig.Method}, existing fairness approaches usually contain two loss term, target classification loss $l_t$ and fairness regularization loss $l_s$. As most fairness research is done under classification tasks, $l_t$ could be either cross-entropy loss or multi-label soft margin loss, optimizing the training data's accuracy. As for the fairness regularization loss $l_s$, it is designed to remove the bias between two sensitive groups. 

In our multi-exit training framework, we duplicate the loss function, $loss=(l_t+ \lambda l_s)$, used in previous work into every internal classifier (IC), $loss^{IC}=(l^{IC}_t+\lambda l^{IC}_s)$, and the final loss is defined by the weighted summation of them, $loss=\alpha_1\cdot loss^{IC1}+\alpha_2\cdot loss^{IC2}+\alpha_3\cdot loss^{IC3}+\alpha_f\cdot loss^{CLS}$, where $\lambda$ is a hyperparameter that controls the trade-off between fairness and accuracy. As both features at shallow and deep layer are included into the loss function, the model will optimize to increase the accuracy and the fairness from shallow to deep naturally.


In addition, as introduced in Section \ref{sec:intro}, since we observe that the fairness would drop at the deeper layer, it is recommended to replace the prediction of the final layer with the internal layer. Based on the heuristic that confidence indicates the correctness of a prediction, we preserve both fairness and accuracy during inference by allowing any instance with high confidence from an internal classifier to exit early. We pre-define a confidence threshold $\theta$, and select the result from the earliest internal classifier in which the confidence is above the threshold. This early exit policy successfully selects the optimal prediction in terms of the trade-off between accuracy and fairness.

\section{Experimental Settings}
\subsection{Datasets}

In this work, we evaluate our framework on two facial attribute datasets, CelebA \cite{liu2018large} and UTK Face. The CelebA dataset consists of over 200k images, each with 40 binary attributes. Similar to \cite{park2022fair}, we set \emph{Male}, and \emph{Young} as the sensitive attributes and select the target attributes which have the highest Pearson correlation with both sensitive attributes \cite{torfason2016face}. In these attributes, we pick \emph{Attractive}, \emph{Big Nose},  and \emph{Bags Under Eyes}. We abbreviate the target attribute (T) and the sensitive attributes (S), \emph{Attractive}, \emph{Big Nose}, \emph{Bags Under Eyes}, \emph{Male}, and \emph{Young} as $a$, $b$, $e$, $m$, and $y$, respectively. 

UTK Face dataset consists of over 20k face images with 3 annotations, \emph{Ethnicity}, \emph{Age}, and \emph{Gender}. We follow the setting in \cite{jung2021fair} to set \emph{Ethnicity} as the sensitive attribute (including \emph{White}, \emph{Black}, \emph{Asian}, and \emph{Indian}.) and divided the \emph{Age} into 3 ranges (ages between 0 to 19, 20 to 40, and larger than 40.) for the target attribute. For FSCL+, we follow the original paper's setting and set the \emph{Gender} as the target attribute.

To show a fair comparison, we follow the recommended setting in previous works to divide both datasets into training/val/test. Results of the test set are reported and discussed in Section \ref{sec:result}.

\subsection{Evaluation Metrics}
\label{sec:evaluation_metrics}
Several fairness metrics are proposed to evaluate the degree of fairness in classification task. Demographic parity \cite{dwork2012fairness} and equalized odds (EO) \cite{hardt2016equality,dwork2012fairness} are two well-known fairness criterion. We first define an input feature $X \in \mathbb{R}^d$ with sensitive attribute $A$, ground truth target class $Y$, and the predicted target class $\hat{Y}$. Demographic parity is satisfied if 
\begin{equation}
\label{eq.DEO}
P(\hat{Y} = 1|A = 0) = P(\hat{Y} = 1|A = 1).
\end{equation}
The drawback of demographic parity is that the classifier could achieve the fairness condition by adjusting the proportion of correct rate of two sensitive attributes through misclassifying some instances. On the other hand, EO forces the true positive rate and the false positive rate along different groups to be equal, that is, 
\begin{equation}
\label{eq.EO}
P(\hat{Y} = 1|A = a, Y = y) = P(\hat{Y} = 1|A = b, Y = y)
\end{equation}
where $y \in \{0, 1\}$, and $a, b \in A$. This metric addresses unfair wrong prediction of the model, and therefore EO will be the suitable fairness metric to evaluate our model. We calculate the degree of EO by calculating the disparity of the $TPR$ and $FPR$ alone different sensitive attributes as follows:
\begin{equation}
\label{eq.EO_equation}
    \sum_{k=1}^{K} |TPR_{k}^{1} - TPR_{k}^{0} + FPR_{k}^{1} - FPR_{k}^{0}|
\end{equation}
where $TPR_{k}^{a}$ and $FPR_{k}^{a}$ are the True Positive Rate and the False Positive Rate respectively of target class $k$ and sensitive attribute $a$, this equation can also extend to multi-attribute cases. In conclusion, the optimal EO score becomes $0$ when the True Positive Rate and the False Positive of each target and sensitive class are the same. It indicates that a lower EO represents the prediction is fairer.


\subsection{Implementation Details}
We utilize Resnet-18 as the backbone of every state-of-the-art and the baseline CNN used in our experiments. 
In the training phase, the data is augmented by random flipping, rotation, and scaling before being fed into the model. The network is trained for 200 epochs using SGD optimizer \cite{ruder2016overview} with an initial learning rate set as 0.01. We attach three internal classifiers at the end of each residual block. The internal classifier contains one adaptive average pooling layer and a two-layer MLP. As introduced in Section \ref{sec:method_main}, the modified loss function in the multi-exit training framework is a weighted summation of the loss from each internal classifier. Since we observe that the learning capacity of shallow ICs are weaker than deep ICs. The coefficients, $\alpha_1$, $\alpha_2$, $\alpha_3$, and $\alpha_f$, are set to $0.3$, $0.45$, $0.6$, and $0.9$, respectively. We tune the confidence threshold based on the model's performance on the validation set and set the confidence threshold $\theta=0.85$ for the early exit policy at the inference phase.  

To show that our scheme can be widely applied to most existing frameworks, we conduct experiments on four baselines selected from state-of-art approaches for fairness-aware learning: LNL \cite{kim2019learning}, HSIC \cite{quadrianto2019discovering}, MFD \cite{jung2021fair}, and FSCL+ \cite{park2022fair}. All the baselines are reproduced by following the recommended hyperparameter settings of the original papers or resources.

\subsubsection{Multi-Exit (ME) Implementation}
We apply our ME framework on four state-of-art methods as below:

\begin{enumerate}
\item ME-CNN : The ME-CNN is our baseline multi-exits framework. We apply cross-entropy loss to each IC without any fairness constraint, that is $loss = \alpha_1\cdot l_t^{IC1}+\alpha_2\cdot l_t^{IC2}+\alpha_3\cdot l_t^{IC3}+\alpha_f\cdot l_t^{CLS}$.

\item ME-LNL : For the adversarial debias method, we replace $g\circ f$ in the original paper with ME-CNN. The loss function becomes $loss = \alpha_1\cdot l_t^{IC1}+\alpha_2\cdot l_t^{IC2}+\alpha_3\cdot l_t^{IC3}+\alpha_f\cdot l_t^{CLS} + \lambda \cdot Adv\_loss$, and the model is trained with an adversarial strategy. $Adv\_loss$ is the adversarial term for bias mitigation. 

\item ME-HSIC : We use ME-CNN as the backbone and apply the proposed loss term in HSIC to each IC.

\item ME-MFD : We follow the teacher-student training framework in the original paper and use ME-CNN as the backbone. We first train a ME-CNN as our teacher model and apply the proposed loss term in MFD to each IC in the student model.

\item ME-FSCL+ : We first pretrain the ICs in a ME-CNN backbone with the fair supervised contrastive loss in FSCL+ together.  Then, we fix the pretrained ME-CNN backbone and train the linear classifiers of each IC together.
\end{enumerate}

\begin{table}[ht]
\centering
\caption{Classification results of the fairness (EO) and accuracy (Acc.) evaluation on the test set of the UTK Face dataset. * denotes our own implementation. T and S stand for target and sensitive attributes, respectively. respectively.}
\label{Table:Compare_SOTA_UTK}

\begin{tabular}{c|c|c|c}
\hline
                              &                             & \multicolumn{2}{c}{S=Ethnicity}         \\ \hline
                              &                             & EO                               & Acc.          \\ \cline{2-4}
\multirow{8}{*}{T=Age}        & CNN*                        & 17.8                             & 82.4          \\ 
                              & \textbf{ME-CNN}             & \textbf{16.0}                    & \textbf{82.5} \\ \cline{2-4}
                              & LNL*                        & 18.9                             & 81.1          \\ 
                              & \textbf{ME-LNL}             & \textbf{16.4}                    & \textbf{81.9} \\ \cline{2-4}
                              & HSIC*                       & 16.4                             & 79.4          \\ 
                              & \textbf{ME-HSIC}            & \textbf{14.9}                    & 78.5          \\ \cline{2-4}
                              & MFD*                        & 15.1                             & 79.1          \\ 
                              & \textbf{ME-MFD}             & \textbf{13.7}                    & \textbf{83.1} \\ \hline
\multirow{2}{*}{T=Gender}     & FSCL+*                      & 3.7                              & 70.1 \\ 
                              & \textbf{ME-FSCL+}           & \textbf{3.2}                     & 70.0  \\ \hline
\end{tabular}

\end{table}



\section{Results}
\label{sec:result}
\begin{figure*}[t]
\begin{center}
\includegraphics[width=0.85\linewidth]{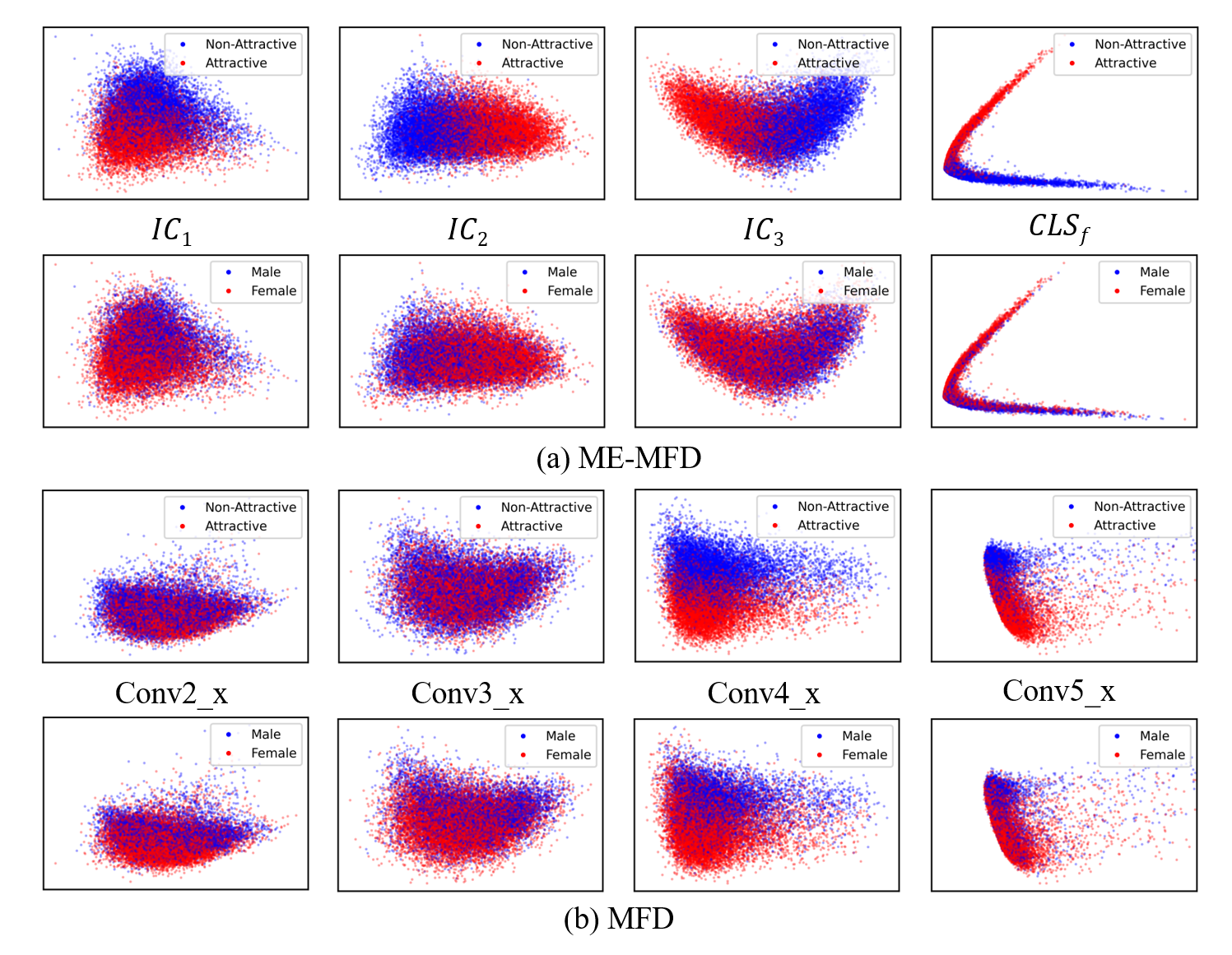}

\caption{Visualization of the features of the CelebA dataset (T=a / S=m). (a) demonstrates the feature of the proposed multi-exit training framework on MFD, while (b) are the features without using the multi-exit training. $IC_x$ represents the $x^{th}$ internal classifier. $CLS_f$ stands for the classifier following behind the last layer. Since the MFD did not contain internal classifier, we extract the feature from the same layers of ME-MFD, which are Conv2\_x, Conv3\_x, Conv4\_x, and Conv5\_x.
}
\label{Fig.ME_ablation}
\end{center}
\end{figure*}
\begin{table*}[ht]
\centering
\caption{Classification results of the proposed ME-MFD and MFD selected from each exit. $IC_x$ represents the $x^{th}$ internal classifier. $CLS_f$ stands for the classifier following behind the last layer. Since the MFD did not contain internal classifier, we extract the feature from the same layers of ME-MFD, which are Conv2\_x, Conv3\_x, Conv4\_x, and Conv5\_x. The fairness (EO) and accuracy (Acc.) evaluation on the test set of the CelebA dataset are reported.}

\setlength{\tabcolsep}{3pt}
\label{Table:Ablation_ME}
\begin{tabular}{c|c||c|c||c|c||c|c||c|c||c|c||c|c}
\hline
\multicolumn{2}{c||}{\multirow{2}{*}{Methods}}                                & \multicolumn{2}{c||}{T=a / S=m} &\multicolumn{2}{c||}{T=a / S=y} &\multicolumn{2}{c||}{T=b / S=m}& \multicolumn{2}{c||}{T=b / S=y}& \multicolumn{2}{c||}{T=e / S=m} & \multicolumn{2}{c}{T=e / S=y}  \\ \cline{3-14}
\multicolumn{2}{c||}{}                                 & EO            & Acc.          & EO            & Acc.          & EO           & Acc.          & EO           & Acc.          & EO           & Acc.          & EO           & Acc.          \\ \hline \hline
\multirow{4}{*}{ME-MFD} & $IC_1$     & \textbf{1.4}  & 70.5          & \textbf{12.9} & 76.9          & \textbf{1.4} & 80.1          & \textbf{3.4} & 78.2          & \textbf{0.8} & 79.6          & \textbf{1.8} & 80.0          \\ 
                        & $IC_2$     & 2.0           & 75.7          & 13.4          & 80.0          & 1.5          & 81.7          & 7.1          & 78.2          & 2.4          & 83.2          & 3.2          & 84.0          \\ 
                        & $IC_3$     & 2.2           & 77.0          & 16.4          & 81.7          & 4.6          & 83.7          & 12.0         & 78.9          & 4.4          & 84.6          & 3.5          & 85.1          \\  
                        & $CLS_f$     & 13.3          & \textbf{79.9} & 23.4          & \textbf{83.0} & 17.7         & \textbf{84.6} & 18.0         & \textbf{79.2} & 12.2         & \textbf{85.3} & 9.6          & \textbf{85.3} \\ \hline
\multirow{4}{*}{MFD}    & Conv2\_x & \textbf{12.2} & 68.9          & \textbf{13.3} & 70.5          & \textbf{4.5} & 78.5          & \textbf{3.7} & 78.7          & \textbf{0.1} & 79.7          & \textbf{1.9} & 79.4          \\ 
                        & Conv3\_x & 16.0          & 73.2  & 15.3          & 75.3          & 19.1         & 80.0          & 10.9         & 79.7          & 8.4          & 80.8          & 3.7          & 80.3          \\  
                        & Conv4\_x & 23.1          & 80.5          & 19.2          & \textbf{79.3}          & 27.6         & 80.0          & 16.9         & 81.4          & 18.7         & 83.1          & 7.8          & 82.8          \\  
                        & Conv5\_x & 23.1          & \textbf{81.3}          & 15.6          & 79.0 & 12.5         & \textbf{80.7} & 8.0          & \textbf{83.0} & 13.1         & \textbf{83.3} & 6.5          & \textbf{83.1} \\ \hline
\end{tabular}

\end{table*}
\begin{table*}[ht]
\centering
\caption{Classification results of the proposed ME-MFD selected from each exit. $IC_x$ represents the $x^{th}$ internal classifier. $CLS_f$ stands for the classifier following behind the last layer. ME-MFD (EE) denotes the result of using the confidence-based early exit algorithm. The fairness (EO) and accuracy (Acc.) evaluation on the test set of the CelebA dataset are reported.}
\setlength{\tabcolsep}{4pt}
\label{Table:Ablation_EE}
\begin{tabular}{c||c|c||c|c||c|c||c|c||c|c||c|c}
\hline
\multirow{2}{*}{Methods}           & \multicolumn{2}{c||}{T=a / S=m} &\multicolumn{2}{c||}{T=a / S=y} &\multicolumn{2}{c||}{T=b / S=m}& \multicolumn{2}{c||}{T=b / S=y}& \multicolumn{2}{c||}{T=e / S=m} & \multicolumn{2}{c}{T=e / S=y}  \\ \cline{2-13}
           & EO           & Acc.           & EO            & Acc.          & EO           & Acc.          & EO           & Acc.           & EO           & Acc.            & EO           & Acc.             \\ \hline \hline
$IC_1$     & \textbf{1.4} & 70.5           & \textbf{12.9} & 76.9          & \textbf{1.4} & 80.1          & \textbf{3.4} & 78.2           & \textbf{0.8} & 79.6            & \textbf{1.8} & 80.0             \\ 
$IC_2$     & 2.0          & 75.7           & 13.4          & 80.0          & 1.5          & 81.7          & 7.1          & 78.2           & 2.4          & 83.2            & 3.2          & 84.0             \\ 
$IC_3$     & 2.2          & 77.0           & 16.4          & 81.7          & 4.6          & 83.7          & 12.0         & 78.9           & 4.4          & 84.6            & 3.5          & 85.1             \\ 
$CLS_f$    & 13.3         & \textbf{79.9}  & 23.4          & \textbf{83.0} & 17.7         & \textbf{84.6} & 18.0         & \textbf{79.2}  & 12.2         & \textbf{85.3}   & 9.6          & \textbf{85.3}    \\ \hline
ME-MFD (EE)& 5.8          & 78.3           & 11.4          & 79.5          & 2.6          & 82.1          & 3.3          & 82.6           & 1.4          & 81.9            & 1.5          & 84.2             \\ \hline
\end{tabular}
\end{table*}

\subsection{Comparison with State-of-the-art}

In this section, we show the feasibility of ME training framework on four state-of-the-art, LNL \cite{kim2019learning}, HSIC \cite{quadrianto2019discovering}, FSCL+ \cite{park2022fair}, and MFD \cite{jung2021fair}, by comparing the result with and without the ME training framework on CelebA and UTK Face dataset in Table \ref{Table:Compare_SOTA_CelebA} and Table \ref{Table:Compare_SOTA_UTK}, respectively.

For the CelebA dataset, in Table \ref{Table:Compare_SOTA_CelebA} we follow the sensitive and target groups setting in \cite{park2022fair} and compare our results with their reproduce results accordingly. The ME-CNN is the baseline ME framework training without any fairness constraint. That is, the loss function in each internal classifier is $loss^{IC} = l_{t}^{IC}$. The ME-CNN framework improves the EO in 20.3\% while losing only 0.13\% of accuracy in average. In the comparison with different state-of-the-art, our results achieve a 38.5\% EO improvement in average while keeping the competitive accuracy. It is noteworthy that in ME-MFD and ME-LNL, our framework also improves the accuracy by an average of 3.8\% and 1.3\% , respectively.

For UTK Face dataset in Table \ref{Table:Compare_SOTA_UTK}, we follow \cite{jung2021fair} to define \emph{Ethnicity} as the sensitive attribute and \emph{Age} as the target groups. Compared with the CNN baseline, the EO decreased from 17.8 to 16, which is a 10\% improvement. As for the comparison with different state-of-the-art, our results achieve a 11.3\% EO improvement in average. In addition, in ME-LNL and ME-MFD, the accuracy also shows a 3\% improvement in average.
Since the FSCL+ \cite{park2022fair} selected the \emph{Gender} as the target attribute, we follow their data imbalance setting to product the experiment. The bias level hyperparameter $N$ is set as 5, which means male data is five times as much as female data, and the other sensitive group has the opposite gender ratio. Performance comparison between FSCL+ and ME-FSCL+ shows that ME-FSCL+ achieves a 13.5\% improvement at EO than FSCL+, where the accuracy is almost the same.
Comparisons in Table \ref{Table:Compare_SOTA_UTK} successfully demonstrate that our framework outperforms all the baseline on the fairness score with a competitive accuracy in UTK Face dataset.

\subsection{Ablation Study of Multi-Exit Training Framework}

In this section, we establish the ablation study of applying a multi-exit training framework to the MFD \cite{jung2021fair}.
In Fig. \ref{Fig.ME_ablation}(a), we visualized the representation of the test instance from each internal classifier.
Since the multi-exit training framework optimized each internal classifier ($IC_x$ and $CLS_f$) for high accuracy and fairness, we can observe a clear decision boundary between the attractive and non-attractive target class. However, the features of the sensitive attribute (gender) are mixed and uniformly distributed, which indicates that the model remains fair.

In addition, we also visualize the representation of the test instance from the same layer of MFD, which is the output of each residual module of the network in Fig. \ref{Fig.ME_ablation}(b). Obviously, since the intermediate feature does not pass through any direct target optimization, both the target and the sensitive attributes could not be easily recognized in shallow layers (Conv2\_x and Conv3\_x). At the deeper layer, the loss function maximizes the accuracy and minimizes the bias of the sensitive group, the feature of the target group can be clearly separated, whereas the sensitive attribute remains the same. We also report the quantitative results of the experiment mentioned above to show the feasibility of the multi-exit training framework.

In Table \ref{Table:Ablation_ME}, we compare the classification results of ME-MFD and MFD selected from each exit branch. We report the features of MFD at different residual modules (i.e., Conv2\_x, Conv3\_x, Conv4\_x, and Conv5\_x) and each IC's features of ME-MFD.
The multi-exit framework improves the EO by an average of 27\% and the accuracy by 1\%.  Besides, the EO of MFD increases from Conv2\_x to Conv\_4 but drops at the last layer. This phenomenon is due to the Maximum Mean Discrepancy (MMD) constraint being imposed at the last layer while there is no fairness constraint on the features of shallow layers. ME-MFD regularizes the fairness condition of each IC; as a result, the trend of EO score still holds our observation in Section \ref{sec:motivation}.

Experiments demonstrate the significance of using the multi-exit training framework, which allows us to select prediction at a shallow layer to achieve a fair and high accuracy result.

\subsection{Ablation Study of Early Exit Policy}

In this section, we study the impact of using the proposed early exit policy. Table \ref{Table:Ablation_EE} shows the fairness and accuracy of the proposed ME-MFD selected from a different exit. 
From the table, we observe that the $IC_1$ obtains the smallest EO and the accuracy, while the $CLS_f$ achieves the largest. The increasing trend in both metrics indicates that there is a trade-off between fairness and accuracy. Thus, the proposed early exit policy independently selects the exit of each test instance to preserve a large fairness improvement with only a slight drop in accuracy.
In the comparison without using the proposed early exit ($CLS_f$), our method achieves 74.5\% lower EO, but the accuracy drop is less than 1.7\% on average. The significant improvement demonstrates the importance of deciding the exit for each instance, which also shows that the proposed early exit algorithm is essential.

\section{Conclusion}
In this paper, we first explored the problem that the fairness condition deteriorates as we classify the features in deeper layers. Then, we introduced the multi-exit training framework, with high extensibility that could be applied to many bias mitigation methods. With the confidence-based exit strategy, we select the optimal exit for each test instance to achieve both high accuracy and fairness. The extensive results and the ablation studies have shown that our framework can achieve the best trade-off of accuracy and fairness conditions compared to the state-of-the-art on two well-known facial datasets.

\bibliography{aaai23}

\begin{thebibliography}{24}
\providecommand{\natexlab}[1]{#1}

\bibitem[{Creager et~al.(2019)Creager, Madras, Jacobsen, Weis, Swersky,
  Pitassi, and Zemel}]{creager2019flexibly}
Creager, E.; Madras, D.; Jacobsen, J.-H.; Weis, M.; Swersky, K.; Pitassi, T.;
  and Zemel, R. 2019.
\newblock Flexibly fair representation learning by disentanglement.
\newblock In \emph{International conference on machine learning}, 1436--1445.
  PMLR.

\bibitem[{Dressel and Farid(2018)}]{dressel2018accuracy}
Dressel, J.; and Farid, H. 2018.
\newblock The accuracy, fairness, and limits of predicting recidivism.
\newblock \emph{Science advances}, 4(1): eaao5580.

\bibitem[{Dwork et~al.(2012)Dwork, Hardt, Pitassi, Reingold, and
  Zemel}]{dwork2012fairness}
Dwork, C.; Hardt, M.; Pitassi, T.; Reingold, O.; and Zemel, R. 2012.
\newblock Fairness through awareness.
\newblock In \emph{Proceedings of the 3rd innovations in theoretical computer
  science conference}, 214--226.

\bibitem[{Gretton et~al.(2012)Gretton, Borgwardt, Rasch, Sch{\"o}lkopf, and
  Smola}]{gretton2012kernel}
Gretton, A.; Borgwardt, K.~M.; Rasch, M.~J.; Sch{\"o}lkopf, B.; and Smola, A.
  2012.
\newblock A kernel two-sample test.
\newblock \emph{The Journal of Machine Learning Research}, 13(1): 723--773.

\bibitem[{Hardt, Price, and Srebro(2016)}]{hardt2016equality}
Hardt, M.; Price, E.; and Srebro, N. 2016.
\newblock Equality of opportunity in supervised learning.
\newblock \emph{Advances in neural information processing systems}, 29.

\bibitem[{Jung et~al.(2021)Jung, Lee, Park, and Moon}]{jung2021fair}
Jung, S.; Lee, D.; Park, T.; and Moon, T. 2021.
\newblock Fair feature distillation for visual recognition.
\newblock In \emph{Proceedings of the IEEE/CVF conference on computer vision
  and pattern recognition}, 12115--12124.

\bibitem[{Kamiran and Calders(2012)}]{kamiran2012data}
Kamiran, F.; and Calders, T. 2012.
\newblock Data preprocessing techniques for classification without
  discrimination.
\newblock \emph{Knowledge and information systems}, 33(1): 1--33.

\bibitem[{Kaya, Hong, and Dumitras(2019)}]{kaya2019shallow}
Kaya, Y.; Hong, S.; and Dumitras, T. 2019.
\newblock Shallow-deep networks: Understanding and mitigating network
  overthinking.
\newblock In \emph{International conference on machine learning}, 3301--3310.
  PMLR.

\bibitem[{Kim et~al.(2019)Kim, Kim, Kim, Kim, and Kim}]{kim2019learning}
Kim, B.; Kim, H.; Kim, K.; Kim, S.; and Kim, J. 2019.
\newblock Learning not to learn: Training deep neural networks with biased
  data.
\newblock In \emph{Proceedings of the IEEE/CVF Conference on Computer Vision
  and Pattern Recognition}, 9012--9020.

\bibitem[{Liu et~al.(2018)Liu, Luo, Wang, and Tang}]{liu2018large}
Liu, Z.; Luo, P.; Wang, X.; and Tang, X. 2018.
\newblock Large-scale celebfaces attributes (celeba) dataset.
\newblock \emph{Retrieved August}, 15(2018): 11.

\bibitem[{Lu et~al.(2020)Lu, Mardziel, Wu, Amancharla, and
  Datta}]{lu2020gender}
Lu, K.; Mardziel, P.; Wu, F.; Amancharla, P.; and Datta, A. 2020.
\newblock Gender bias in neural natural language processing.
\newblock In \emph{Logic, Language, and Security}, 189--202. Springer.

\bibitem[{Mehrabi et~al.(2021)Mehrabi, Morstatter, Saxena, Lerman, and
  Galstyan}]{mehrabi2021survey}
Mehrabi, N.; Morstatter, F.; Saxena, N.; Lerman, K.; and Galstyan, A. 2021.
\newblock A survey on bias and fairness in machine learning.
\newblock \emph{ACM Computing Surveys (CSUR)}, 54(6): 1--35.

\bibitem[{Ngxande, Tapamo, and Burke(2020)}]{ngxande2020bias}
Ngxande, M.; Tapamo, J.-R.; and Burke, M. 2020.
\newblock Bias remediation in driver drowsiness detection systems using
  generative adversarial networks.
\newblock \emph{IEEE Access}, 8: 55592--55601.

\bibitem[{Park et~al.(2022)Park, Lee, Lee, Hwang, Kim, and Byun}]{park2022fair}
Park, S.; Lee, J.; Lee, P.; Hwang, S.; Kim, D.; and Byun, H. 2022.
\newblock Fair Contrastive Learning for Facial Attribute Classification.
\newblock In \emph{Proceedings of the IEEE/CVF Conference on Computer Vision
  and Pattern Recognition}, 10389--10398.

\bibitem[{Quadrianto, Sharmanska, and Thomas(2019)}]{quadrianto2019discovering}
Quadrianto, N.; Sharmanska, V.; and Thomas, O. 2019.
\newblock Discovering fair representations in the data domain.
\newblock In \emph{Proceedings of the IEEE/CVF conference on computer vision
  and pattern recognition}, 8227--8236.

\bibitem[{Ruder(2016)}]{ruder2016overview}
Ruder, S. 2016.
\newblock An overview of gradient descent optimization algorithms.
\newblock \emph{arXiv preprint arXiv:1609.04747}.

\bibitem[{Schwartz et~al.(2020)Schwartz, Stanovsky, Swayamdipta, Dodge, and
  Smith}]{schwartz2020right}
Schwartz, R.; Stanovsky, G.; Swayamdipta, S.; Dodge, J.; and Smith, N.~A. 2020.
\newblock The right tool for the job: Matching model and instance complexities.
\newblock \emph{arXiv preprint arXiv:2004.07453}.

\bibitem[{Teerapittayanon, McDanel, and
  Kung(2016)}]{teerapittayanon2016branchynet}
Teerapittayanon, S.; McDanel, B.; and Kung, H.-T. 2016.
\newblock Branchynet: Fast inference via early exiting from deep neural
  networks.
\newblock In \emph{2016 23rd International Conference on Pattern Recognition
  (ICPR)}, 2464--2469. IEEE.

\bibitem[{Torfason et~al.(2016)Torfason, Agustsson, Rothe, and
  Timofte}]{torfason2016face}
Torfason, R.; Agustsson, E.; Rothe, R.; and Timofte, R. 2016.
\newblock From face images and attributes to attributes.
\newblock In \emph{Asian Conference on Computer Vision}, 313--329. Springer.

\bibitem[{Wang et~al.(2022)Wang, Dong, Xue, Zhang, Chiu, Wei, and
  Ren}]{wang2022fairness}
Wang, Z.; Dong, X.; Xue, H.; Zhang, Z.; Chiu, W.; Wei, T.; and Ren, K. 2022.
\newblock Fairness-aware Adversarial Perturbation Towards Bias Mitigation for
  Deployed Deep Models.
\newblock In \emph{Proceedings of the IEEE/CVF Conference on Computer Vision
  and Pattern Recognition}, 10379--10388.

\bibitem[{Zhang, Lemoine, and Mitchell(2018)}]{zhang2018mitigating}
Zhang, B.~H.; Lemoine, B.; and Mitchell, M. 2018.
\newblock Mitigating unwanted biases with adversarial learning.
\newblock In \emph{Proceedings of the 2018 AAAI/ACM Conference on AI, Ethics,
  and Society}, 335--340.

\bibitem[{Zhang, Song, and Qi(2017)}]{zhang2017age}
Zhang, Z.; Song, Y.; and Qi, H. 2017.
\newblock Age progression/regression by conditional adversarial autoencoder.
\newblock In \emph{Proceedings of the IEEE conference on computer vision and
  pattern recognition}, 5810--5818.

\bibitem[{Zhao et~al.(2017)Zhao, Wang, Yatskar, Ordonez, and
  Chang}]{zhao2017men}
Zhao, J.; Wang, T.; Yatskar, M.; Ordonez, V.; and Chang, K.-W. 2017.
\newblock Men also like shopping: Reducing gender bias amplification using
  corpus-level constraints.
\newblock \emph{arXiv preprint arXiv:1707.09457}.

\bibitem[{Zhou et~al.(2020)Zhou, Xu, Ge, McAuley, Xu, and Wei}]{zhou2020bert}
Zhou, W.; Xu, C.; Ge, T.; McAuley, J.; Xu, K.; and Wei, F. 2020.
\newblock Bert loses patience: Fast and robust inference with early exit.
\newblock \emph{Advances in Neural Information Processing Systems}, 33:
  18330--18341.

\end{thebibliography}

\end{document}